\documentclass{article}

\usepackage[T1]{fontenc}
\usepackage{amsmath}
\usepackage{amsthm}
\usepackage{mathrsfs}
\usepackage{amssymb}
\usepackage{algorithm}
\usepackage{algorithmic}
\newtheorem{theorem}{Theorem}
\newtheorem{proposition}{Proposition}

\newtheorem{lemma}{Lemma}
\newtheorem{definition}{Definition}
\usepackage{wrapfig}
\usepackage[utf8x]{inputenc} \DeclareMathOperator*{\argmax}{arg\,max}

\def\bs#1{\boldsymbol{#1}}

\def\nopo{P(\tilde{\bs{Y}}|X = \bs{x})}
\def\clpo{P(\bs{Y}|X = \bs{x})}

\def\nopos{P(\tilde{\bs{Y}}|X)}
\def\clpos{P(\bs{Y}|X)}

\usepackage{graphicx}
\usepackage{subfigure}
\usepackage{booktabs} 

\usepackage{hyperref}

\usepackage[accepted]{icml2021}

\icmltitlerunning{Provably End-to-end Label-noise Learning without Anchor Points}

\begin{document}

\twocolumn[
\icmltitle{Provably End-to-end Label-noise Learning without Anchor Points}



\icmlsetsymbol{equal}{*}

\begin{icmlauthorlist}
\icmlauthor{Xuefeng Li}{1,2}
\icmlauthor{Tongliang Liu}{2}
\icmlauthor{Bo Han}{3}
\icmlauthor{Gang Niu}{4}
\icmlauthor{Masashi Sugiyama}{4,5}
\end{icmlauthorlist}

\icmlaffiliation{1}{University of New South Wales}
\icmlaffiliation{2}{Trustworthy Machine Learning Lab, University of Sydney}
\icmlaffiliation{3}{Hong Kong Baptist University}
\icmlaffiliation{4}{RIKEN AIP}
\icmlaffiliation{5}{University of Tokyo}

\icmlcorrespondingauthor{Tongliang Liu}{tongliang.liu@sydney.edu.au}

\icmlkeywords{Machine Learning, ICML}

\vskip 0.3in
]



\printAffiliationsAndNotice{} 

\begin{abstract}
In label-noise learning, the \emph{transition matrix} plays a key role in building \emph{statistically consistent} classifiers. Existing consistent estimators for the transition matrix have been developed by exploiting \textit{anchor points}. However, the anchor-point assumption is not always satisfied in real scenarios. In this paper, we propose an end-to-end framework for solving label-noise learning without anchor points, in which we simultaneously optimize two objectives: the cross entropy loss between the noisy label and the predicted probability by the neural network, and the volume of the simplex formed by the columns of the transition matrix.~Our proposed framework can identify the transition matrix if the clean class-posterior probabilities are \emph{sufficiently scattered}. This is by far the mildest assumption under which the transition matrix is provably \textit{identifiable} and the learned classifier is statistically consistent. Experimental results on benchmark datasets demonstrate the effectiveness and robustness of the proposed method.

\end{abstract}

\section{Introduction} \label{sec1}

The success of modern deep learning algorithms heavily relies on large-scale accurately annotated data \cite{daniely2019generalization, han2020survey, xia2020part, berthon2020confidence}. However, it is often expensive or even infeasible to annotate large datasets. Therefore, cheap but less accurate annotating methods have been widely used \cite{xiao2015learning, li2017webvision, han2020sigua, yu2020label, zhu2020second}. As a consequence, these alternatives inevitably introduce label noise. Training deep learning models on noisy data can significantly degenerate the test performance due to overfitting to the noisy labels \cite{arpit2017closer, zhang2016understanding, xia2021robust, wu2020class2simi}.


To mitigate the negative impacts of label noise, many methods have been developed and some of them are based on a loss correction procedure. In general, these methods are \textit{statistically consistent}, i.e., these methods guarantee that the classifier learned from the noisy data approaches to the optimal classifier defined on the clean risk as the size of the noisy training set increases \cite{liu2016classification,scott2015rate,natarajan2013learning,goldberger2016training,patrini2017making,thekumparampil2018robustness}. The idea is that the clean class-posterior $\clpo :=[P(Y=1|X=\bs{x}),\ldots, P(Y=C|X=\bs{x})]^\top$ can be inferred by utilizing the noisy class-posterior $\nopo$ and the transition
matrix $\bs{T}(\bs{x})$ where $\bs{T}_{ij}(\bs{x}) = P(\tilde{Y}=i | Y=j, X = \bs{x})$, i.e., $\clpo = [\bs{T}(\bs{x})]^{-1} \nopo$. While those methods  theoretically guarantee the statistical consistency, they all heavily rely on the success of estimating transition matrices.

Generally, the transition matrix is unidentifiable without additional assumptions \cite{xia2019anchor}. In the literature, methods have been developed to estimate the transition matrices under the so-called \emph{anchor-point} assumption: it assumes the existence of anchor points, i.e., instances belonging to a specific class with probability one \cite{liu2016classification}. The assumption is reasonable in certain applications \cite{liu2016classification,patrini2017making}. However, the violation of the assumption in some cases could lead to a poorly learned transition matrix and a degenerated classifier \cite{xia2019anchor}. This motivates the development of algorithms without exploiting anchor points \cite{xia2019anchor, liu2020peer, xu2019l_dmi, zhu2021clusterability}. However, the performance is not theoretically guaranteed in these works.

\begin{figure}
\centering
\includegraphics[width=0.4\textwidth]{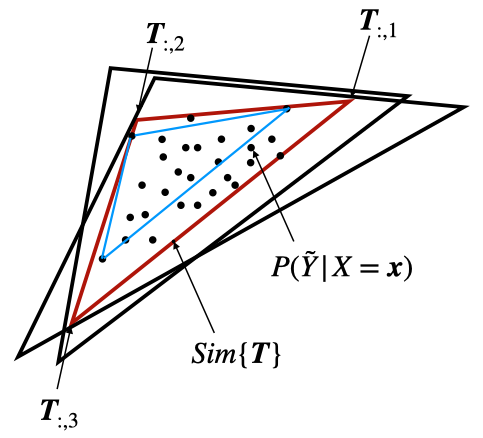}
\caption{Geometric illustration of the problem of estimating the transition matrix without anchor points. The red triangle is the simplex of $\bs{T}$ with vertices denoted by $\bs{T}_{:;i}$. When there are no anchor points, the simplex found with existing methods (blue triangle) by using extreme-valued noisy class-posterior probabilities (see Eq. (\ref{anchor})) is not the true simplex (red triangle). It is obvious that among possible enclosing simplexes (black and red triangles), the true simplex $\mathrm{Sim}\{\bs{T}\}$ has the minimum volume.}
\label{volmin_intuition}
\vspace{-14pt}
\end{figure}

{\bf{Motivation}}. In this work, our interest lies in designing a consistent algorithm without anchor points, subject to class-dependent label noise, i.e., $\bs{T(\bs{x})} = \bs{T}$ for any $\bs{x}$ in the feature space. Our algorithm is based on a geometric property of the label corruption process. Given an instance $\bs{x}$, the noisy class-posterior probability $\nopo :=[P(\tilde{Y}=1|X=\bs{x}),\ldots, P(\tilde{Y}=C|X=\bs{x})]^\top$  can be thought of as a point in the $C$-dimensional space where $C$ is the number of classes. Since we have $\nopo = \bs{T} \clpo$ and $\sum_{i=1}^{C} P(Y=i|X = \bs{x}) = 1$, $\nopo$ is then a convex combination of the columns of $\bs{T}$. This means that the simplex $\mathrm{Sim}\{\bs{T}\}$ formed by the columns of $\bs{T}$ encloses $\nopo$ for any $\bs{x}$ \cite{boyd2004convex}. Thus, the problem of identifying the transition matrix can be treated as the problem of recovering $\mathrm{Sim}\{\bs{T}\}$. However, when no assumption has been made, the problem is ill-defined as $\mathrm{Sim}\{\bs{T}\}$ is not identifiable, i.e., there exists an infinite number of simplexes enclosing $\nopos$, and any of them can be regarded as the true simplex $\mathrm{Sim}\{\bs{T}\}$. It is apparent that under the anchor-point assumption, $\mathrm{Sim}\{\bs{T}\}$ can be uniquely determined by exploiting anchor points whose noisy class-posterior probabilities are the vertices of $\mathrm{Sim}\{\bs{T}\}$. The goal is thus to identify the points which have the largest noisy class-posterior probabilities for each class \cite{liu2016classification, patrini2017making}. However, if there are no anchor points, the identified points would not be the vertices of $\mathrm{Sim}\{\bs{T}\}$. In this case, existing methods cannot consistently estimate the transition matrices. To recover $\bs{T}$ without anchor points, a key observation is that, among all simplexes enclosing $\nopos$, $\mathrm{Sim}\{\bs{T}\}$ is the one with \emph{minimum volume}. See Figure \ref{volmin_intuition} for a geometric illustration. This observation motivates the development of our method which incorporates the minimum volume constraint of $\mathrm{Sim}\{\bs{T}\}$ into label-noise learning.

To this end, we propose \textit{Volume Minimization Network} (VolMinNet) to consistently estimate the transition matrix and build a statistically consistent classifier. Specifically, VolMinNet consists of a classification network $\bs{h}_{\bs{\theta}}$ and a trainable transition matrix $\hat{\bs{T}}$.  We simultaneously optimize $\hat{\bs{T}}$ and $\bs{h}_{\bs{\theta}}$ with two objectives: i) the discrepancy between $\hat{\bs{T}}\bs{h}_{\bs{\theta}}(\bs{x})$ and the noisy class-posterior distribution $\nopo$, ii) The volume of the simplex formed by the columns of $\hat{\bs{T}}$. The proposed framework is end-to-end, and there is no need for identifying anchor points or \emph{pseudo anchor points} (i.e., instances belonging to a specific class with probability close to one) \cite{xia2019anchor}. Since our proposed method does not rely on any specific data points, it yields better noise robustness compared with existing methods. With a so-called \emph{sufficiently scattered assumption} where the clean class-posterior distribution is far from uniform, we theoretically prove that $\hat{\bs{T}}$ will converge to the true transition matrix $\bs{T}$ while $\bs{h}_{\bs{\theta}}(\bs{x})$ converges to the clean class-posterior $\clpo$. We also prove that the anchor-point assumption is a special case of the sufficiently scattered assumption. 

The rest of this paper is organized as follows. In Section 2, we set up the notations and review the background of label-noise learning with anchor points. In Section 3, we introduce our proposed VolMinNet. In Section 4, we present the main theoretical results. In Section 5, we briefly introduce the related works in the literature. Experimental results on both synthetic and real-world datasets are provided in Section 6. Finally, we conclude the paper in Section 7.

\section{Label-Noise Learning with Anchor Points}

%
 
In this section, we review the background of label-noise learning. We follow common notational conventions in the literature of label-noise learning. $\bs{v} \in \mathbb{R}^{n}$ and $\bs{V} \in \mathbb{R}^{n \times m}$ denote a real-valued $n$-dimensional vector and a real-valued $n \times m$ matrix, respectively. Elements of a vector are denoted by a subscript (e.g., $\bs{v}_{j}$), while rows and columns of a matrix are denoted by $ \bs{V}_{i, :}$ and $\bs{V}_{:, i}$ respectively. The $i$th standard basis vector in $\mathbb{R}^{C}$ is denoted by $\bs{e}_i$. We denote the all-ones vector by $\bs{1}$, and $\Delta^{C-1} \subset[0,1]^{C}$ is the $C$-dimensional simplex. In this work, we also make extensive use of convex analysis. Let a set $\mathcal{V} = \{\bs{v}_{1}, \ldots, \bs{v}_{m}\},$ and the convex cone of $\mathcal{V}$ is denoted by $\mathrm{cone}(\mathcal{V})=\{\bs{v}|\bs{v}=\sum_{j=1}^{m} \bs{v}_{j} \alpha_{j}, \; \alpha_{j} \geq 0, \; \forall  j \}$. Similarly, the convex hull of $\mathcal{V}$ is defined as  $\mathrm{conv}(\mathcal{V})=\{\bs{v}|\bs{v}=\sum_{j=1}^{m} \bs{v}_{j} \alpha_{j}, \; \alpha_{j} \geq 0, \; \sum_{j=1}^{m}\alpha_j = 1, \; \forall  j \}$. Specially, when $\{\bs{v}_{1}, \ldots, \bs{v}_{m}\}$ are affinely independent, $\mathrm{conv}(\mathcal{V})$ is also called a simplex which we denote it as $\mathrm{Sim}(\mathcal{V})$.


Let $\mathcal{D}$  be the underlying distribution generating a pair of random variables $(X, Y) \in \mathcal{X} \times \mathcal{Y},$ where $\mathcal{X} \subseteq \mathbb{R}^d$ is the feature space, $\mathcal{Y} = \{1, 2,\ldots, C\}$ is the label space and $C$ is the number of classes. In many real-world applications, samples drawn from $\mathcal{D}$ are unavailable. Before being observed, labels of these samples are contaminated with noise and we obtain a set of corrupted data $\{(\bs{x}_{i}, \tilde{y}_{i})\}_{i=1}^{n}$ where $\tilde{y}$ is the noisy label and we denote by $\tilde{\mathcal{D}}$ the distribution of the noisy random pair $(X, \tilde{Y})\in \mathcal{X} \times \mathcal{Y}$.


Given an instance $\bs{x}$ sampled from $X$, $\tilde{Y}$ is  derived from the random variable $Y$ through a noise transition matrix $\bs{T}(\bs{x})\in [0, 1]^{C\times C}$:
\begin{equation}
	 \nopo= \bs{T}(x) \clpo,
\end{equation}
where $\clpo=[P(Y=1|X=\bs{x}),\ldots, P(Y=C|X=\bs{x})]^\top$ and $\nopo=[P(\tilde{Y}=1|X=\bs{x}),\ldots, P(\tilde{Y}=C|X=\bs{x})]^\top$ are the clean class-posterior probability and the noisy class-posterior probability, respectively. The $ij$-th entry of the transition matrix, i.e., $\bs{T}_{ij}(\bs{x})=P(\tilde{Y}=i|Y=j, X=\bs{x})$, represents the probability that the instance $\bs{x}$ with the clean label $Y=j$ will have a noisy label $\tilde{Y}=i$. Generally, the transition matrix is \textit{non-identifiable} without any additional assumption \cite{xia2019anchor}. For example, we can decompose the transition matrix with $\bs{T}(\bs{x})=\bs{T}_1(\bs{x})\bs{T}_2(\bs{x})$. If we define $P'(\bs{Y}|X=\bs{x}) = \bs{T}_2(\bs{x}) \clpo$, then $\nopo = \bs{T}(x) \clpo) = \bs{T}_1(\bs{x}) P'(\bs{Y}|X=\bs{x})$ are both valid. Therefore, in this paper, we study the \textit{class-dependent} and \textit{instance-independent} transition matrix on which the majority of existing methods focus \cite{han2018co,han2018masking,patrini2017making,northcuttlearning,natarajan2013learning}. Formally, we have:
\begin{equation}
	\nopo= \bs{T} \clpo,
	\label{trans2}
\end{equation}
where the transition matrix $\bs{T}$ is now independent of the instance $\bs{x}$. In this work, we also assume that the true transition matrix $\bs{T}$ is \textit{diagonally dominant}\footnote{The definition of being diagonally dominant is different from the one in matrix analysis, but it has been commonly used in label-noise learning \cite{xu2019l_dmi}}. Specifically, the transition matrix $\bs{T}$ is diagonally dominant if for every column of $\bs{T}$, the
magnitude of the diagonal entry is larger than any non-diagonal entry, i.e.,  $\bs{T}_{ii} > \bs{T}_{ji}$ for any $j \neq i$. This assumption has been commonly used in the literature of label-noise learning \cite{patrini2017making, xia2019anchor, yao2020dual}.

As in Eq.~(\ref{trans2}), the clean class-posterior probability $\clpos$ can be inferred by using the noisy class-posterior probability $\nopos$ and the transition matrix $\bs{T}$ as $\clpos=\bs{T}^{-1}\nopos$. For this reason, the transition matrix has been widely exploited to build statistically consistent classifiers, i.e., the learned classifier will converge to the optimal classifier defined with clean risk. Specifically, the transition matrix has been used to modify loss functions to build risk-consistent estimators \cite{goldberger2016training, patrini2017making,yu2018learning,xia2019anchor}, and has been used to correct hypotheses to build classifier-consistent algorithms \cite{natarajan2013learning, scott2015rate, patrini2017making}. Thus, the successes of these consistent algorithms rely on an accurately learned transition matrix. 

In recent years, considerable efforts have been invested in designing algorithms for estimating the transition matrix. These algorithms rely on a so-called \textit{anchor-point} assumption which requires that there exist anchor points for each class \cite{liu2016classification,xia2019anchor}. 


\begin{definition}[anchor-point assumption]
	For each class $j \in \{1,2,\ldots,C\}$, there exists an instance $\bs{x}^j \in \mathcal{X}$ such that $P(Y=j|X=\bs{x}^j)=1$.  \end{definition}

 Under the anchor-point assumption, the task of estimating the transition matrix boils down to finding anchor points for each class. For example, given anchor points $\bs{x}^j$, we have 

\begin{equation}
	P(\tilde{Y}=i \mid X=\bs{x}^j)=\sum_{k=1}^{C} T_{i k} P(Y=k \mid X=\bs{x}^j)=T_{i j}.
	\label{est} 
\end{equation}
Namely, the transition matrix can be obtained with the noisy class-posterior probabilities of anchor points. Assuming that we can accurately model the noisy class-posterior $\nopos$ given a sufficient number of noisy data, anchor points can be easily found as follows \cite{liu2016classification, patrini2017making}: 
\begin{equation}
	\bs{x}^j= \argmax_{\bs{x}}P(\tilde{Y}=j|X=\bs{x}).
	\label{anchor}
\end{equation}
However, when the anchor-point assumption is not satisfied, points found with Eq. (\ref{anchor}) are no longer anchor points. Hence, the above-mentioned method can not consistently estimate the transition matrix with Eq. (\ref{est}), which will lead to a statistically inconsistent classifier. This motivates us to design a statistically classifier-consistent algorithm which can consistently estimate the transition matrix without anchor points. 
 

\section{Volume Minimization Network}

\begin{figure*}[ht]
\begin{center}
	\includegraphics[width=150mm,height=45mm]{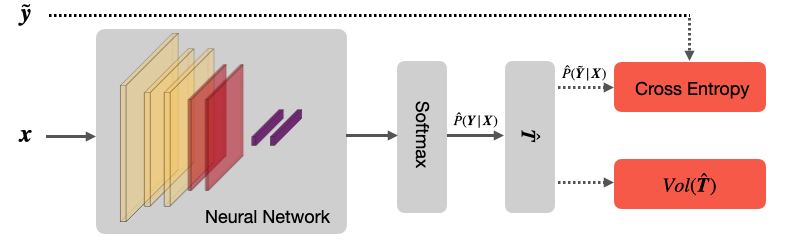}
\end{center}
\caption{Overview of the proposed VolMinNet. The training in the proposed framework is carried out in an end-to-end manner with two objectives (red blocks) optimized simultaneously.}
\label{overview}
\end{figure*}

In this section, we propose a novel framework for label-noise learning which we call the Volume Minimization Network (VolMinNet). The proposed framework is end-to-end, and there is no need for identifying anchor points or a second stage for loss correction, resulting in better noise robustness than existing methods.

To learn the clean class-posterior $P(Y|X)$, we define a transformation $h_{\bs{\theta}}: \mathcal{X} \rightarrow \Delta^{C-1}$ where $h_{\bs{\theta}}$ is a differentiable function represented by a neural network with parameters $\bs{\theta}$. To estimate the transition matrix, we construct a trainable \textit{diagonally dominant column stochastic} matrix $\hat{\bs{T}}$, i.e., $\hat{\bs{T}} \in[0,1]^{C \times C}$, $\sum_{i=1}^{C} \bs{T}_{ij} = 1$  and $\bs{T}_{ii} > \bs{T}_{ji}$ for any $i \neq j$. To learn the noisy class posterior distribution from the noisy data, with some abuse of notation, we define the composition of $\hat{\bs{T}}$ and $h_{\bs{\theta}}$ as $\hat{\bs{T}}h_{\bs{\theta}} : \mathcal{X} \rightarrow \Delta^{C-1}$.

 

Intuitively, as explained in Section \ref{sec1}, if $\hat{\bs{T}}h_{\bs{\theta}}$ models $\nopos$ perfectly while the simplex of $\hat{\bs{T}}$ has the minimum volume, $\hat{\bs{T}}$ will converge to the true transition matrix $\bs{T}$ and $h_{\bs{\theta}}$ will converge to $\clpos$. This motivates us to propose the following criterion which corresponds to a constraint optimization problem:


\begin{equation}
\begin{aligned}
	\min _{\hat{\bs{T}} \in \mathbb{T}} & \;\; \mathrm{vol}(\hat{\bs{T}}) \\
\text { s.t. } & \hat{\bs{T}}h_{\bs{\theta}}= \nopos,
\end{aligned}
\label{cons_p}
\end{equation}
where $\mathbb{T} =\{\hat{\bs{T}} \in [0,1]^{C \times C} \; | \; \sum_{i=1}^{C} \hat{\bs{T}}_{ij} = 1, \; \hat{\bs{T}}_{ii} > \bs{T}_{ji}, \; \forall \; i \neq j\}$ is the set of diagonally dominant column stochastic matrices. $\mathrm{vol}(\hat{\bs{T}})$ denotes a measure that is related or proportional to the volume of the simplex formed by the columns of  $\hat{\bs{T}}$.

To solve criterion (\ref{cons_p}), we first note that the constraint $\hat{\bs{T}}h_{\bs{\theta}}=\nopos$ can be solved with expected risk minimization \cite{patrini2017making}. The risk is defined as $\tilde{R}(h_{\bs{\theta}})=\mathbb{E}_{(\bs{x}, \tilde{y}) \sim \tilde{D}}[\ell(\hat{\bs{T}}h_{\bs{\theta}}(\bs{x}), \tilde{y})]$, where $\ell$ is a loss function and we use the cross-entropy loss throughout this paper. We can then re-write criterion (\ref{cons_p}) as a Lagrangian under the KKT condition \cite{karush1939minima, kuhn2014nonlinear} to obtain:
\begin{equation}
	\mathcal{L}(\theta, \hat{\bs{T}}):= \mathrm{vol}(\hat{\bs{T}}) + \beta \cdot \mathbb{E}_{(\bs{x}, \tilde{y}) \sim \tilde{D}}[\ell(\hat{\bs{T}}h_{\bs{\theta}}(\bs{x}), \tilde{y})],
\end{equation}
where $\beta>0$ is the KKT multiplier. In the literature, various functions for measuring the volume of the simplex have been investigated \cite{fu2015blind, li2008minimum, miao2007endmember}. Given $\hat{\bs{T}}$ is a square matrix, a common choice is $\mathrm{vol}(\hat{\bs{T}}) = \mathrm{det}(\hat{\bs{T}})$, where $\mathrm{det}$ denotes the determinant. However, this function is numerically unstable for optimization and computationally hard to deal with. Hence, we adopt another popular alternative $\log \mathrm{det}(\hat{\bs{T}})$. This function has been widely used in low-rank matrix recovery and non-negative matrix decomposition \cite{fazel2003log, liu2012robust, fu2016robust}. Besides, since we only have access to a set of noisy training examples $\{(\bs{x}_{i}, \tilde{y}_{i})\}_{i=1}^{n}$ instead of the distribution $\tilde{\mathcal{D}}$, we employ the empirical risk for training. Formally, we propose the following objective function: 
\begin{equation}
\label{obj}
	\mathcal{L}(\theta, \hat{\bs{T}}):= \frac{1}{n} \sum_{i=1}^{n} \tilde{\ell}(\hat{\bs{T}}h_{\bs{\theta}}(\bs{x}_i)), \tilde{y}_{i})  + \lambda \cdot \log \mathrm{det}(\hat{\bs{T}}),
\end{equation}
where $\lambda > 0$ is a regularization coefficient that balances distribution fidelity versus volume minimization. 

The problem remains how to design $\hat{\bs{T}}$ so that it is differentiable, diagonally dominant and column stochastic. Specifically, we first create a matrix $\bs{A} \in \mathbb{R}^{C \times C}$ so that diagonal elements $\bs{A}_{ii} = 1$ for all $i \in \{1,2,\ldots,C\}$, and all other elements $\bs{A}_{ij} = \sigma(w_{ij})$ for all $i \neq j$ where $\sigma$ is the sigmoid function $\sigma(w) = \frac{1}{1+e^{-w}}$ and each $w_{ij} \in \mathbb{R}$ is a real-valued variable which will be updated throughout training. Then we do the normalization $\hat{\bs{T}}_{ij}= \frac{\bs{A}_{ij}}{\sum_{k=1}^{C} \bs{A}_{kj}}$ so that the sum of each column of $\hat{\bs{T}}$ is equal to one. Since the sigmoid function returns a value  in the range 0 to 1, we have $\hat{\bs{T}}_{ii} > \hat{\bs{T}}_{ji}$ for all $i  \neq j$. With this specially designed $\hat{\bs{T}}$, we ensure that $\hat{\bs{T}} \in \mathbb{T}$, i.e., $\hat{\bs{T}}$ is a diagonally dominant and column stochastic matrix. In addition, $\hat{\bs{T}}$ is differentiable because the sigmoid function and the normalization operation are differentiable.

With both $\hat{\bs{T}}$ and $h_{\bs{\theta}}$ being differentiable, the objective in Eq. (\ref{obj}) can be easily optimized with any standard gradient-based learning rule. This allows us to replace the two-stage loss correction procedure in existing works with an end-to-end learning framework. See Figure \ref{overview} for a less formal, more pedagogical explanation of our proposed learning framework.

\section{Theoretical Results}

In this section, we show that criterion (\ref{cons_p}) guarantees the consistency of the estimated transition matrix and the learned classifier under the sufficiently scattered assumption. We also show that the anchor-point assumption is a special case of the sufficiently assumption. To explain, we give a formal definition of the sufficiently scattered assumption:

\begin{figure*}[ht]
\begin{center}
	\includegraphics[width=150mm,height=41mm]{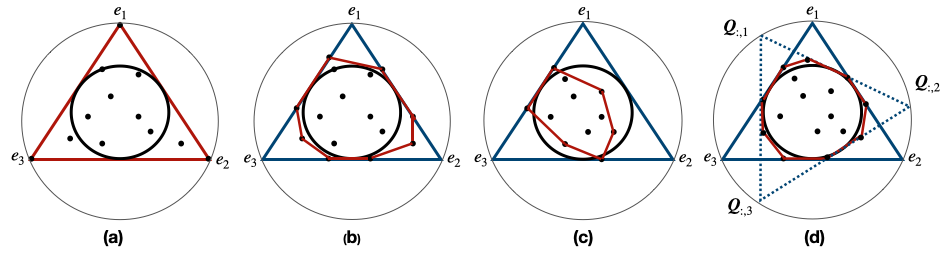}
\end{center} 
	\caption{Illustration of the anchor-point assumption and the sufficiently scattered assumption in the case of $C = 3$ by assuming that the viewer are facing the hyperplane $\bs{1}^{\top}\bs{x} = 1$ from the positive orthant. The dots are class-posterior probabilities $\clpo$; the triangle is the non-negative orthant; the inner circle is $\mathcal{R}$; the region encompassed by red lines is $\mathrm{cone}\{\bs{H}\}$. Clearly, the anchor-point assumption (a) is a special case of the sufficiently scattered assumption (b).}
	\label{sc}
\end{figure*}
\begin{definition}[Sufficiently Scattered] The clean class-posterior $\clpos$ is said to be sufficiently scattered if there exists a set $\mathcal{H} = \{x_1, \ldots, x_m\} \subseteq \mathcal{X}$ such that the matrix $\bs{H} = [P(\bs{Y}|X=x_1), \ldots ,P(\bs{Y}|X=x_m)]$ satisfies (1) $\mathcal{R} \subseteq \mathrm{cone}\{\bs{H}\},$ \; where $\mathcal{R}=\{\boldsymbol{v} \in \mathbb{R}^C \mid \boldsymbol{v}^{\top} \boldsymbol{1} \geq \sqrt{C-1}\|\boldsymbol{v}\|_{2}\}$ and $\mathrm{cone}\{\bs{H}\}$ denotes the convex cone formed by columns of $\bs{H}$. (2) $\mathrm{cone}\{\bs{H}\} \nsubseteq cone\{\bs{Q}\}$, \;for any unitary matrix $\bs{Q} \in \mathbb{R}^{C\times C}$ that is not a permutation matrix. 
	\label{def:sc}
\end{definition}

This assumption is evolved from previous works in non-negative matrix decomposition \cite{fu2015blind, fu2018identifiability} with necessary modification. Intuitively, in the case of $C = 3$, $\mathcal{R}$ corresponds to a ``ball'' tangential to the triangle formed by a permutation matrix, e.g., $\bs{I}=[\bs{e}_1,\bs{e}_2,\bs{e}_3]$. $\mathrm{cone}\{\bs{H}\}$ is the polytope inside this triangle. Columns of $\bs{Q}$ also form triangles which are rotated versions of the triangle defined by permutation matrices; facets of those triangles are also tangential to $\mathcal{R}$. Condition (1) of the sufficiently scattered assumption requires that $\mathcal{R}$ is enclosed by $\mathrm{cone}\{\bs{H}\}$, i.e., $\mathcal{R}$ is a subset of $\mathrm{cone}\{\bs{H}\}$. Condition (2) ensures that given condition (1), $\mathrm{cone}\{\bs{H}\}$ is enclosed by the triangle formed by a permutation matrix and not any other unitary matrix. 

To understand the sufficiently scattered assumption and its relationship with the anchor-point assumption, we provide several examples in Figure \ref{sc}.  In Figure \ref{sc}.(a), we show a situation where both the anchor-point assumption and sufficiently scattered assumption are satisfied. Figure \ref{sc}.(b) shows a situation where the sufficiently scattered assumption is satisfied while the anchor-point assumption is violated. In Figure \ref{sc}.(c) and \ref{sc}.(d), both assumptions are violated. However, in Figure \ref{sc}.(d), only condition (2) of the sufficiently scattered assumption is violated while both conditions of the sufficiently scattered assumption are violated in \ref{sc}.(c).

The first observation is that if the anchor-point assumption is satisfied, then the sufficiently scattered assumption must hold, but not vice versa. Intuitively, if the anchor-point assumption is satisfied, then there exists a matrix $\bs{H} = [P(\bs{Y}|X=\bs{x}^1), \ldots, P(\bs{Y}|X=\bs{x}^C)] = \bs{I}$ where $\bs{x}^1, \ldots, \bs{x}^C$ are anchor points for different classes and $\bs{I}$ is the identity matrix. From Figure \ref{sc}.(a) , it is clear that $\mathcal{R} \subseteq  \mathrm{cone}\{\bs{I}\} = \mathrm{cone}\{\bs{H}\},$ and $\mathrm{cone}\{\bs{H}\}$ can only be enclosed by the convex cone of permutation matrices. This shows that the sufficiently scattered assumption is satisfied. However, from Figure \ref{sc}.(b), it is clear that the sufficiently scattered assumption is satisfied but not the anchor-point assumption.  Formally, we show that:

\begin{proposition}
	The anchor-point assumption is a sufficient but not necessary condition for the sufficiently scattered assumption when $C > 2$.
	\label{prop}
\end{proposition}


The proof of Proposition \ref{prop} is included in the supplementary
material. Proposition \ref{prop} implies that the anchor-point assumption is a special case of the sufficiently scattered assumption. This means that the proposed framework can also deal with the case where the anchor-point assumption holds. Under the sufficiently scattered assumption, we get our main result:
\begin{theorem}\label{t2}
Given sufficiently many noisy data, if $\clpos$ is sufficiently scattered, then $\hat{\bs{T}}_{\star}=\bs{T} $ and $h_{\bs{\theta}_{\star}}(\bs{x})=\clpo $ must hold, where $(\hat{\bs{T}}_{\star}, \bs{\theta}_{\star})$ are optimal solutions of Eq. (\ref{cons_p}).
\label{theorem1}
\end{theorem} 

The proof of Theorem \ref{theorem1} can be found in the supplementary
material. Intuitively, if $\clpos$ is sufficiently scattered, the noisy class-posterior $\nopos$ will be sufficiently spread in the simplex formed by the columns of $\bs{T}$. 
Then, finding the minimum-volume data-enclosing convex hull of $\nopos$ recovers the ground-truth $\bs{T}$ and $\clpos$. 


\section{Related Works}
In this section, we review existing methods in label-noise learning. Based on the statistical consistency of the learned classifier, we divided
exsisting methods for label-noise learning into two categories: heuristic algorithms and statistically consistent algorithms. 

Methods in the first category focus on employing heuristics to reduce the side-effect of noisy labels. For example, many methods use a specially designed strategy to select reliable samples \cite{yu2019does,han2018co,malach2017decoupling,ren2018learning,jiang2018mentornet, yao2020searching} or correct labels \cite{ma2018dimensionality,kremer2018robust,tanaka2018joint,reed2014training}. Although those methods empirically work well, there is not any theoretical guarantee on the consistency of the learned classifiers from all these methods.


Statistically consistent algorithms are primarily developed based on a loss correction procedure \cite{ liu2016classification, patrini2017making, zhang2018generalized}. For these methods, the noise transition matrix plays a key role in building consistent classifiers. For example, Patrini et al.\yrcite{patrini2017making} leveraged a two-stage training procedure of first estimating the noise transition matrix and then use it to modify the loss to ensure risk consistency. These works rely on anchor points or instances belonging to a specific class with probability one or approximately one. When there are no anchor points in datasets or data distributions, all the aforementioned methods cannot guarantee the statistical consistency. Another approach is to jointly learn the noise transition matrix and classifier. For instance, on top of the softmax layer of the classification network \cite{goldberger2016training}, a constrained linear layer or a nonlinear softmax layer is added to model the noise transition matrix \cite{sukhbaatar2015training}. Zhang et al. \yrcite{zhang2021confidence} concurrently propose a one-step method for the label-noise learning problem and a derivative-free method for estimating the transition matrix. Specifically, their method uses a total variation regularization term to prevent the overconfidence problem of the neural network, which leads to a more accurate noisy class-posterior. However, the anchor-point assumption is still needed for their method. Based on different motivations, assumptions and learning objectives, their method achieves different theoretical results compared with our proposed method. Learning with noisy labels are also closely related to learning with complementary labels where instead of noisy labels, only compelementary labels are given for training \cite{yu2018learning, chou2020unbiased, feng2020learning}. 

Recently, some methods exploiting semi-supervised learning techniques have been proposed to solve the label-noise learning problem like SELF \cite{nguyen2019self} and DivideMix \cite{li2019dividemix}. These methods are aggregations of multiple techniques such as augmentations and multiple networks. Noise robustness is significantly improved with these methods. However, these methods are sensitive to the choice of hyperparameters and changes in data and noise types would generate degenerated classifiers. In addition, the computational cost of these methods increases significantly compared with previous methods.

\section{Experiments}

   \begin{figure*}
     \centering
     
     \includegraphics[width=0.24\textwidth]{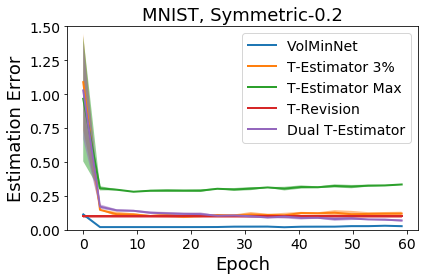}
     \includegraphics[width=0.24\textwidth]{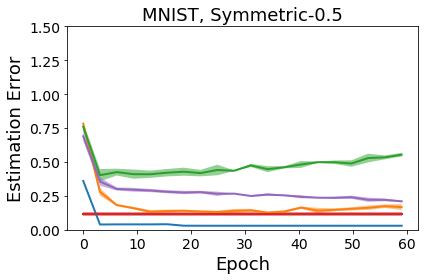}   
     \includegraphics[width=0.24\textwidth]{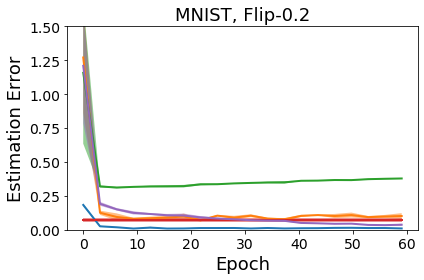}
     \includegraphics[width=0.24\textwidth]{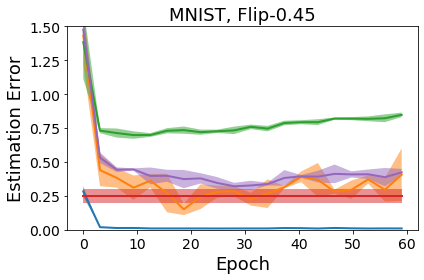}
\vspace{5mm}

     \includegraphics[width=0.24\textwidth]{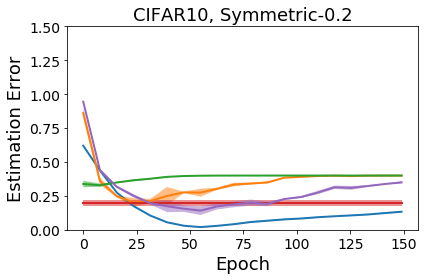}
     \includegraphics[width=0.24\textwidth]{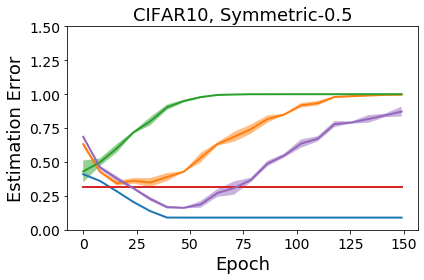}
     \includegraphics[width=0.24\textwidth]{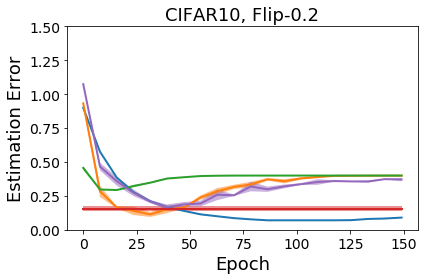}
     \includegraphics[width=0.24\textwidth]{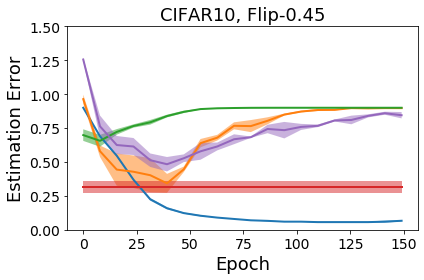}
     
   \vspace{5mm}
  
     \includegraphics[width=0.24\textwidth]{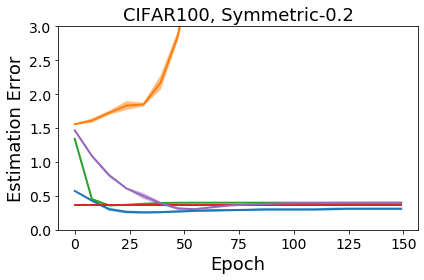}
     \includegraphics[width=0.24\textwidth]{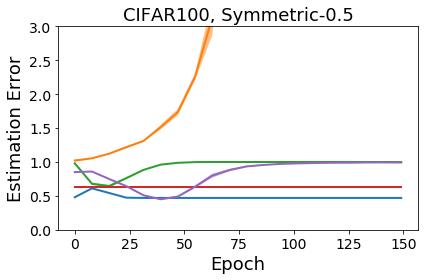}
     \includegraphics[width=0.24\textwidth]{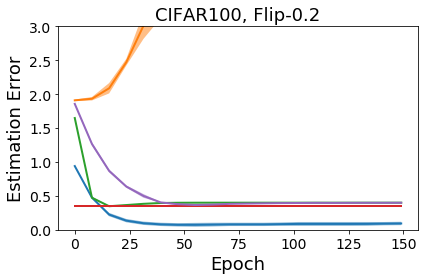}
     \includegraphics[width=0.24\textwidth]{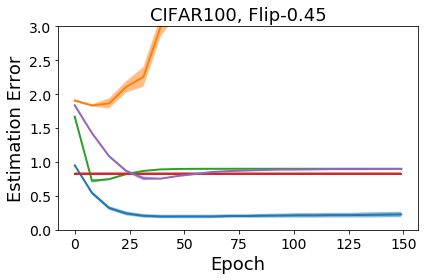}

\vspace{5mm}

     \caption{Transition matrix estimation error on MNIST, CIFAR-10 and CIFAR-100. The error bar for the standard deviation in each figure has been shaded. The lower the better.}
     \label{fig:est_error}
 \end{figure*}
 
In this section, we verify the robustness of the proposed volume minimization network (VolMinNet) from two folds: the estimation error of the transition matrix and the classification accuracy.

\textbf{Datasets} We evaluate the proposed method on three synthetic noisy datasets, i.e.,
MNIST, CIFAR-10 and CIFAR-100 and one real-world noisy dataset, i.e., clothing1M. We leave out 10\% of the training examples as the validation set. The three synthetic datasets contain clean data. We corrupted the training and validation sets manually according to transition matrices. Specifically, we conduct experiments with two commonly used types of noise: (1) Symmetry flipping \cite{patrini2017making}; (2) Pair flipping  \cite{han2018co}. We report both the classification accuracy on the test set and the estimation error between the estimated transition matrix $\hat{\bs{T}}$ and the true transition matrix $\bs{T}$. All experiments are repeated five times on all datasets. Following T-Revision \cite{xia2019anchor}, we also conducted experiments on datasets where possible anchor points are removed from the datasets. The details and more experimental results can be found in the Supplementary Material.


Clothing1M is a real-world noisy dataset which consists of $1 \mathrm{M}$ images with real-world noisy labels. Existing methods like Forward \cite{patrini2017making} and T-revision \cite{xia2019anchor} use the additional 50k clean training data to help initialize the transition matrix and validate on 14k clean validation data. Here we use another setting which is also commonly used in the literature \cite{xia2020part}. We only exploit the $1 \mathrm{M}$ data for both transition matrix estimation and classification training. Specifically, we leave out $10\%$ of the noisy training examples as a noisy validation set for model selection. We think this setting is more natural considering that it does not require any clean data. All results of baseline methods are quoted from PTD \cite{xia2020part} as we have the same setting. 

\textbf{Network structure and optimization} For a fair comparison, we implement all methods with default parameters by PyTorch on Tesla V100-SXM2. For MNIST, we use a LeNet-5 network. SGD is used to train the classification network $h_{\bs{\theta}}$ with batch size $128$, momentum $0.9$, weight decay $10^{−3}$ and a learning rate $10^{-2}$. Adam with default parameters is used to train the transition matrix $\hat{\bs{T}}$. The algorithm is run for $60$ epoch. For CIFAR10, we use a ResNet-18 network. SGD is used to train both the classification network $h_{\bs{\theta}}$ and the transition matrix $\hat{\bs{T}}$ with batch size $128$, momentum $0.9$, weight decay $10^{−3}$ and an initial learning rate $10^{-2}$. The algorithm is run for $150$ epoch and the learning rate is divided by $10$ after the $30$th and $60$th epoch. For CIFAR100, we use a ResNet-32 network. SGD is used to train the classification network $h_{\bs{\theta}}$ with batch size $128$, momentum $0.9$, weight decay $10^{−3}$ and an initial learning rate $10^{-2}$. Adam with default parameters is used to train the transition matrix $\hat{\bs{T}}$. The algorithm is run for $150$ epoch and the learning rate is divided by $10$ after the $30$th and $60$th epoch. For CIFAR-10 and CIFAR-100, we perform data augmentation by horizontal random flips and $32\times 32$ random crops after padding 4 pixels on each side. For clothing1M, we use a ResNet-50 pre-trained on ImageNet. We only use the 1M noisy data to train and validate the network. For the optimization, SGD is used train both the classification network $h_{\bs{\theta}}$ and the transition matrix $\hat{\bs{T}}$ with momentum 0.9, weight decay $10^{−3}$, batch size 32, and run with learning rates $2 \times 10^{−3}$ and $2 \times 10^{−5}$ for 5 epochs each. For each epoch, we ensure the noisy labels for each class are balanced with undersampling. Throughout all experiments, we fixed $\lambda = 10^{-4}$ and the trainable weights $w$ of $\hat{\bs{T}}$ are initialized with $ln\frac{1}{C-2}$ (roughly -2 for MNIST and CIFAR10, -4.5 for CIFAR100 and -2.5 for clothing1M).

\begin{table*}
    \centering
    \scalebox{1}{
\begin{tabular}{ccccccc}

\hline  
& \multicolumn{2}{c} {MNIST} & \multicolumn{2}{c} {CIFAR-10} & \multicolumn{2}{c} {CIFAR-100} \\
& Sym-20\% & Sym-50\% & Sym-20\% & Sym-50\% & Sym-20\% & Sym-50\% \\
\hline
Decoupling & $ 97.04 \pm 0.06$ & $ 94.58 \pm 0.08$ & $ 77.32 \pm 0.35$ & $ 54.07 \pm 0.46$ & $ 41.92 \pm 0.49$ & $ 22.63 \pm 0.44$ \\
MentorNet & $ 97.21 \pm 0.06$ & $ 95.56 \pm 0.15$ & $ 81.35 \pm 0.23$ & $ 73.47 \pm 0.15$ & $ 42.88 \pm 0.41$ & $ 32.66 \pm 0.40$ \\
Co-teaching & $ 97.07 \pm 0.10$ & $ 95.20 \pm 0.23$ & $ 82.27 \pm 0.07$ & $ 75.55 \pm 0.07$ & $ 48.48 \pm 0.66$ & $ 36.77 \pm 0.52$ \\
Forward & $98.60 \pm 0.19$ & $97.77 \pm 0.16$ & $85.20 \pm 0.80$ & $74.82 \pm 0.78$ & $54.90 \pm 0.74$ &$41.85 \pm 0.71$  \\
T-Revision & $ 98.72 \pm 0.10$ & $ \mathbf{98.23 \pm 0.10}$ & $ 87.95 \pm 0.36$ & $ 80.01 \pm 0.62$ & $ 62.72 \pm 0.69$ & $ 49.12 \pm 0.22$ \\
DMI  & $ 98.70 \pm 0.02$ & $ 98.12 \pm 0.21$ & $ 87.54 \pm 0.20$ & $ 82.68 \pm 0.21$ & $ 62.65 \pm 0.39$ & $ 52.42 \pm 0.64$\\
Dual T & $ 98.43 \pm 0.05$ & $ 98.15 \pm 0.12$ & $ 88.35 \pm 0.33$ & $ 82.54 \pm 0.19$ & $ 62.16 \pm 0.58$ & $ 52.49 \pm 0.37$\\
VolMinNet & $\mathbf{9 8 . 7 4} \pm \mathbf{0 . 0 8}$ & $\mathbf{9 8 . 2 3} \pm \mathbf{0 . 1 6}$ & $\mathbf{8 9 . 5 8} \pm \mathbf{0 . 2 6}$ & $\mathbf{8 3 . 3 7} \pm \mathbf{0 . 2 5}$ & $\mathbf{64.94} \pm \mathbf{0.40}$& $\mathbf{53.89} \pm \mathbf{1.26}$ \\

\hline
\end{tabular}
}
    \centering
    \scalebox{1}{
\begin{tabular}{ccccccc}

& \multicolumn{2}{c} {MNIST} & \multicolumn{2}{c} {CIFAR-10} & \multicolumn{2}{c} {CIFAR-100} \\
& Pair-20\% & Pair-45\% &Pair-20\% & Pair-45\% & Pair-20\%& Pair-45\% \\
\hline
Decoupling & $ 96.93 \pm 0.07$ & $ 94.34 \pm 0.54$ & $ 77.12 \pm 0.30$ & $ 53.71 \pm 0.99$ & $ 40.12 \pm 0.26$ & $ 27.97 \pm 0.12$ \\
MentorNet & $ 96.89 \pm 0.04$ & $ 91.98 \pm 0.46$ & $ 77.42 \pm 0.00$ & $ 61.03 \pm 0.20$ & $ 39.22 \pm 0.47$ & $ 26.48 \pm 0.37$ \\
Co-teaching & $ 97.00 \pm 0.06$ & $ 96.25 \pm 0.01$ & $ 80.65 \pm 0.20$ & $ 73.02 \pm 0.23$ & $ 42.79 \pm 0.79$ & $ 27.97 \pm 0.20$ \\
Forward & $98.84 \pm 0.10$ & $75.06 \pm 12.61$ & $88.21 \pm 0.48$ & $77.44 \pm 6.89$ & $56.12 \pm 0.54$ & $36.88 \pm 2.32$ \\
T-Revision & $ 98.89 \pm 0.08$ & $ 84.56 \pm 8.18$ & $ 90.33 \pm 0.52$ & $ 78.94 \pm 2.58$ & $ 64.33 \pm 0.49$ & $ 41.55 \pm 0.95$\\

DMI  & $ 98.84 \pm 0.09$ & $ 97.92 \pm 0.76$ & $ 89.89 \pm 0.45$ & $ 73.15 \pm 7.31$ & $ 59.56 \pm 0.73$ & $ 38.17 \pm 2.02$\\
Dual T & $ 98.86 \pm 0.04$ & $ 96.71 \pm 0.12$ & $ 89.77 \pm 0.25$ & $ 76.53 \pm 2.51$ & $ 67.21 \pm 0.43$ & $ 47.60 \pm 0.43$\\
VolMinNet &$\mathbf{9 9 . 0 1} \pm \mathbf{0 . 0 7}$   & $\mathbf{9 9 . 0 0} \pm \mathbf{0 . 0 7}$ & $\mathbf{9 0 . 3 7} \pm \mathbf{0 . 3 0}$ & $\mathbf{8 8 . 5 4} \pm \mathbf{0 . 2  1}$ & $\mathbf{68 . 45} \pm \mathbf{0 . 6 9}$ &$\mathbf{58 . 9 0} \pm \mathbf{0 . 8 9}$  \\
\hline
\end{tabular}
}

	\caption{Classification accuracy (percentage) on MNIST, CIFAR-10 and CIFAR-100.}
	\label{table:accs}
\end{table*}

\subsection{Transition Matrix Estimation} \label{exp:est}
For evaluating the effectiveness of estimating the transition matrix, we compare the proposed method with the following methods: (1) T-estimator max \cite{patrini2017making}, which identify the extreme-valued noisy class-posterior probabilities from given samples to estimate the transition matrix. (2) T-estimator 3\% \cite{patrini2017making}, which takes a $\alpha$-percentile in place of the argmax of Equation \ref{anchor}.
(3) T-Revision \cite{xia2019anchor}, which introduces a slack variable to revise the noise transition matrix after initializing the transition matrix with T-estimator. (4) Dual T-estimator \cite{yao2020dual}, which introduces an intermediate class to avoid directly estimating the noisy class-posterior and factorizes the transition matrix into the product of two easy-to-estimate transition matrices.

To show that the proposed method is more robust in estimating the transition matrix, we plot the estimation error for the transition matrix, i.e., $\|T-\hat{T}-\Delta \hat{T}\|_{1} /\|T\|_{1}$. Figure~\ref{fig:est_error} depicts estimation errors of transition matrices estimated by the proposed VolMinNet and other baseline methods. For all different settings of noise on three different datasets (original intact datasets), VolMinNet consistently gives better results compared to the baselines, which shows its superior robustness against label noise. For example, on CIFAR100 (Flip-0.45), our method achieves estimation error around 0.25, while baseline methods can only reach at around 0.75. These results show that our method establishes the new state of the art in estimating transition matrices.

\subsection{Classification accuracy Evaluation}

We compare the classification accuracy of the proposed method with the following methods: (1) Decoupling \cite{malach2017decoupling}. (2) MentorNet \cite{jiang2018mentornet}. (3) Co-teaching \cite{han2018co}. (4) Forward \cite{patrini2017making}. (5) T-Revision \cite{xia2019anchor}. (7) DMI \cite{xu2019l_dmi}. (8) Dual T \cite{yao2020dual}. Note that we did not compare the proposed method with some methods like SELF \cite{nguyen2019self} and DivideMix \cite{li2019dividemix}. This is because these methods are aggregations of semi-supervised learning techniques which have high computational complexity and are sensitive to the choice of hyperparameters. In this work, we are more focusing on solving the label noise learning without anchor points theoretically.

In Table~\ref{table:accs}, we present the classification accuracy by the proposed VolMinNet and baseline methods on synthetic noisy datasets. VolMinNet outperforms baseline methods on almost all settings of noise. This result is natural after we have shown that VolMinNet leads to smaller estimation error of the transition matrix compared with baseline methods. While the differences of accuracy among different methods are marginal for symmetric noise,  
VolMinNet outperforms baselines by over $10\%$ with Pair-$45\%$ noise and has much smaller standard deviations. These results show the clear advantage of the proposed VolMinNet. It has better robustness to different settings of noise and datasets compared to baseline methods.

\begin{table}
\centering
$
\begin{array}{cccc}
\hline \text { Decoupling } & \text { MentorNet } & \text { Co-teaching } & \text { Forward } \\
\hline  54.53 & 56.79 & 60.15 & 69.91\\

\hline
\text { T-Revision } &  \text { DMI } &\text { PTD } &\text { VolMinNet } \\
\hline 70.97 & 70.12 & 71.67 & \mathbf{72.42}\\

\hline
\end{array}	
$
\caption{Classification accuracy (percentage) on Clothing1M. Only noisy data are exploited for training and validation.}
\label{clothing}
\end{table}

Finally, we show the results on Clothing1M in Table \ref{clothing}. As explained in the previous section,  Forward and T-Revision exploited the 50k clean data and their noisy versions in 1M noisy data to help initialize the noise transition matrix, which is not practical in real-world settings. For a fair comparison, we report results by only using the $1M$ noisy data to train and validate the network. As shown in Table \ref{clothing}, our method outperforms previous transition matrix based methods and heuristic methods on the Clothing1M dataset. In addition, the performance on the Clothing1M dataset shows that the proposed method has certain robustness against instance-dependent noise as well.

\section{Discussion and Conclusion}

In this paper, we considered the problem of label-noise learning without anchor points. We relax the anchor-point assumption with our proposed VolMinNet. The consistency of the estimated transition matrix and the learned classifier are theoretically proved under the sufficiently scattered assumption. Experimental results have demonstrated the robustness of the proposed VolMinNet. Future work should focus on improving the estimation of the noisy class posterior which we believe is the bottleneck of our method. 

\section*{Acknowledgements}
XL was supported by an Australian Government RTP Scholarship. TL was supported by Australian Research Council Project DE-190101473. BH was supported by the RGC Early Career Scheme No. 22200720, NSFC Young Scientists Fund No. 62006202 and HKBU CSD Departmental Incentive Grant. GN and MS were supported by JST AIP Acceleration Research Grant Number JPMJCR20U3, Japan. MS was also supported by Institute for AI and Beyond, UTokyo. Authors also thank for the help from Dr. Alan Blair, Kevin Lam, Yivan Zhang and members of the Trustworthy Machine Learning Lab at the University of Sydney.


\bibliography{icml}
\bibliographystyle{icml2021}

\clearpage

\appendix
\section{Appendix}
\subsection{Proof of Proposition 1}
To prove proposition 1, we first show that the anchor-point assumption is a sufficient condition for the sufficiently scattered assumption. In other words, we need to show that if the anchor-point assumption is satisfied, then two conditions of the sufficiently scattered assumption must hold.

We start with condition (2) of the sufficiently scattered assumption. We need to show that if the anchor-point assumption is hold, then there exists a set $\mathcal{H} = \{x_1, \ldots, x_m\} \subseteq \mathcal{X}$ such that the matrix $\bs{H} = [P(Y|X=x_1), \ldots ,P(Y|X=x_m)]$ satisfies that $\mathrm{cone}\{\bs{H}\} \nsubseteq \mathrm{cone}\{\bs{Q}\}$, \;for any unitary matrix $\bs{Q} \in \mathbb{R}^{C\times C}$ that is not a permutation matrix. 

Since the anchor-point assumption is satisfied, then there exists a matrix $\bs{H} = [P(Y|X=\bs{x}^1),..., P(Y|X=\bs{x}^C)]$  where $\bs{x}^1, ..., \bs{x}^C$ are anchor points for each class. From the definition of anchor points, we have $P(Y|X=\bs{x}^i) = \bs{e}_i$. This implies that
 
\begin{equation}
	\bs{H} = [P(Y|X=\bs{x}^1), ..., P(Y|X=\bs{x}^C)] = \bs{I},
	\label{prop_eq:1}
\end{equation}

where $\bs{I}$ is the identity matrix. By the definition of the identity matrix $\bs{I}$, it is clear that $\mathrm{cone}\{\bs{H}\} = \mathrm{cone}\{\bs{I}\} \nsubseteq \mathrm{cone}\{\bs{Q}\}$, \;for any unitary matrix $\bs{Q} \in \mathbb{R}^{C\times C}$ that is not a permutation matrix. This shows that condition (2) of the sufficiently scattered assumption is satisfied if the anchor-point assumption is hold.

Next, we show that condition (1) will also be satisfied, i.e., the convex cone $\mathcal{R} \subseteq \mathrm{cone}\{\bs{H}\},$ where $\mathcal{R}=\{\boldsymbol{v} \in \mathbb{R}^C | \boldsymbol{v}^{\top} \boldsymbol{1} \geq \sqrt{C-1}\|\boldsymbol{v}\|_{2}\}$. By Eq. (\ref{prop_eq:1}), condition (1) of Theorem 1 is equivalent to 

\begin{equation}
	\mathcal{R}  \subseteq \mathrm{cone}\{\bs{I}\} =\{\bs{u}|\bs{u}=\sum_{j=1}^{C} \bs{e}_{j} \alpha_{j}, \; \alpha_{j} \geq 0, \; \forall  j \}.
\end{equation}

This means that all elements in $\mathcal{R}$ must be in the non-negative orthant of $\mathbb{R}^C$, i.e., for all $\bs{v} \in \mathcal{R}$, $\bs{v}_i \geq 0$ for all $i \in \{1,\ldots,C\}$. Consider $\bs{v} \in \mathcal{R}$ and let $\hat{\bs{v}}$ be the normalized vector of $\bs{v}$, by definition of $\mathcal{R}$ we have the following chain:

\begin{subequations}
	\begin{align}
	\boldsymbol{v}^{\top} \boldsymbol{1} &\geq \sqrt{C-1}\|\boldsymbol{v}\|_{2},\\
	\frac{\boldsymbol{v}^{\top}}{\|\boldsymbol{v}\|} \boldsymbol{1} = \hat{\bs{v}}^{\top}\bs{1} &\geq \sqrt{C-1},\\
	\sum_{i \in \{1,2,\ldots,C\}} \hat{\bs{v}}_i &\geq \sqrt{C-1}.
\end{align}
\label{equation:10}
\end{subequations}

To show $\bs{v}$ is non-negative is equivalent to prove that $\hat{\bs{v}}$ is non-negative, i.e.,  $\forall k \in \{1,\ldots,C\}$, $\hat{\bs{v}}_k \geq 0$. Let $\bs{u} \in \mathbb{R}^{C-1}$ be the vector which has same elements with $\hat{\bs{v}}$ except that the $k$th element $\hat{\bs{v}}_k$ is removed. Following Eq. \ref{equation:10}, we have:

\begin{subequations}
\begin{align}
	\;\hat{\bs{v}}_k &\geq \sqrt{C-1} - \sum_{i \in \{1,2,\ldots,C\} \setminus \{k\}}\hat{\bs{v}}_i,\\
	\hat{\bs{v}}_k &\geq \sqrt{C-1} - \bs{u}^{\top}\bs{1}.
\end{align}
\label{equation:11}
\end{subequations}

By the Cauchy-Schwarz inequality, we get the following inequality:

\begin{equation}
	|\bs{u}^{\top}\bs{1}| \leq \Vert \bs{u}\Vert \Vert \bs{1} \Vert.
	\label{equation:12}
\end{equation}

Then by the definition of $\bs{u}$ and $\bs{1}$, we have $\Vert \bs{u}\Vert \leq 1$ and $\Vert \bs{1} \Vert = \sqrt{C-1}$. Combined this with Eq. \ref{equation:11} and Eq. \ref{equation:12}, we get:

\begin{equation}
		\hat{\bs{v}}_k \geq \sqrt{C-1} - \Vert \bs{u}\Vert \Vert \bs{1} \Vert \geq 0.
\end{equation}

This simply implies that $\bs{v}_k \geq 0$ for all $k \in \{1,2,\ldots,C\}$ and we have proved that the anchor-point assumption is a sufficient condition of the sufficiently scattered assumption. 

We now prove that the anchor-point assumption is not a necessary condition for the sufficiently scattered assumption. Suppose $\clpos$ has the property that $x^1, x^2, \ldots x^C \notin \mathcal{X}$ which means that the anchor-point assumption is not satisfied. We also assume that there exist a set $\mathcal{H} = \{x_1, \ldots, x_m\} \subseteq \mathcal{X}$ such that $\mathrm{cone}\{\bs{H}\}$ covers the whole non-negative orthant except the area along each axis (area formed by noisy class-posterior of anchor points). Since these areas along each axis are not part of $\mathcal{R}$ when $C > 2$, it is clear that condition (1) of the sufficiently scattered assumption is satisfied. Besides, by definition of $\bs{H}$, there is no other unitary matrix which can cover $\mathrm{cone}\{\bs{H}\}$ except permutation matrices. This shows that condition (2) of the sufficiently scattered assumption is also satisfied and the proof is completed. \qed

\subsection{Proof of Theorem 1}
The insights of our proof are from previous works in non-negative matrix factorisation \cite{fu2015blind}. To proceed, let us first introduce following classic lemmas in convex analysis:

\begin{lemma}
If $\mathcal{K}_{1}$ and $\mathcal{K}_{2}$ are convex cones and $\mathcal{K}_{1} \subseteq \mathcal{K}_{2},$ then, $dual\{\mathcal{K}_{2}\} \subseteq dual\{\mathcal{K}_{1}\}$. 
\label{lemma1}
\end{lemma}

\begin{lemma} 
If $\bs{A}$ is invertible, then $dual(\mathbf{A})=\mathrm{cone}(\mathbf{A}^{-\top})$.
\label{lemma2}
\end{lemma}

Readers are referred to Boyd et al.\yrcite{boyd2004convex} for details.
Our purpose is to show that criterion (5) has unique solutions which are the ground-truth $\clpos$ and $\bs{T}$. To this end, let us denote $(\bs{T}_\star, h_{\bs{\theta}_\star})$ as a feasible solution of Criterion (5), i.e.,

\begin{equation}
	\bs{T}_{\star}h_{\bs{\theta}_\star}= \bs{T}\clpos = \nopos.
	\label{eq.8}
\end{equation}

As defined in sufficient scattered assumption, we have the matrix $\bs{H} = [P(\bs{Y}|X=x_1), \ldots ,P(\bs{Y}|X=x_m)]$ defined on the set $\mathcal{H} = \{x_1, \ldots, x_m\} \subseteq \mathcal{X}$. Let $\bs{H}_{\star} = [h_{\bs{\theta}_\star}(\bs{x}_1), \ldots ,h_{\bs{\theta}_\star}(\bs{x}_m)]$, it follows that
\begin{equation}
	\bs{T}_{\star}\bs{H}_{\star}= \bs{T}\bs{H}.
\label{eq:9}
\end{equation}

Note that both $\bs{T}_{\star}$ and $\bs{T}$ have full rank because they are diagonally dominant square matrices by definition. In addition, since the sufficiently scattered assumption is satisfied, $rank(\bs{H}) = C$ also holds \cite{fu2015blind}. Therefore, there exists an invertible matrix $\bs{A} \in \mathbb{R}^{C \times C}$ such that
\begin{equation}
\bs{T}_{\star} = \bs{T}\bs{A}^{-\top},
\label{eq:10}
\end{equation}

where $\bs{A}^{-\top} = \bs{H}\bs{H}_{\star}^{\dagger}$ and  $\bs{H}_{\star}^{\dagger} = \bs{H}_{\star}^{\top}(\bs{H}_{\star}\bs{H}_{\star}^{\top})^{-1}$ is the pseudo-inverse of $\bs{H}_{\star}$.

Since $\bs{1}^{\top} \bs{H}=\bs{1}^{\top}$ and $\mathbf{1}^{\top} \bs{H}_{\star}=\mathbf{1}^{\top}$ by definition, we get
\begin{equation}
	\mathbf{1}^{\top} \bs{A}^{-\top}=\mathbf{1}^{\top} \bs{H} \bs{H}_{\star}^{\dagger}=\mathbf{1}^{\top} \bs{H}_{\star}^{\dagger}=\mathbf{1}^{\top} \bs{H}_{\star} \bs{H}_{\star}^{\dagger}=\mathbf{1}^{\top}.
	\label{eq.11}
\end{equation}

Let $\bs{v} \in \mathrm{cone}\{\bs{H}\},$ which by definition takes the form $\bs{v}=\bs{H} \bs{u}$
for some $\bs{u} \geq 0 .$ Using $\bs{H}=\bs{A}^{-\top} \bs{H}_{\star}, \; \bs{v}$ can be expressed as $\bs{v} = \bs{A}^{-\top} \bs{\tilde{u}}$ where $\bs{\tilde{u}}=\bs{H}_{\star} \bs{u} \geq 0$. This implies that $\bs{v}$ also lies in $\mathrm{cone} \{\bs{A}^{-\top}\}$, i.e. $\mathrm{cone}\{\bs{H}\} \subseteq \mathrm{cone}\{\bs{A}^{-\top}\}$.

Recall Condition (1) of the sufficiently scattered assumption, i.e., $\mathcal{R} \subseteq \mathrm{cone}\{\bs{H}\}$ where 
$\mathcal{R}=\{\bs{v} \in \mathbb{R}^{C} | \mathbf{1}^{\top} \bs{v} \geq \sqrt{C-1}\|\bs{v}\|_{2}\} .$ It
implies
\begin{equation}
\mathcal{R} \subseteq \mathrm{cone}\{\bs{H}\} \subseteq \mathrm{cone}(\bs{A}^{-\top}).
\label{eq.12}
\end{equation}

By applying Lemmas (\ref{lemma1}-\ref{lemma2}) to Eq. (\ref{eq.12}), we have
\begin{equation}
	\mathrm{cone}(\bs{A}) \subseteq dual\{\mathcal{R}\},
	\label{13}
\end{equation}
where $dual\{\mathcal{R}\}$ is the dual cone of $\mathcal{R},$ which can be shown to be

\begin{equation}
	dual\{\mathcal{R}\}=\{\bs{v} \in \mathbb{R}^{C} | \|\bs{v}\|_{2} \leq \mathbf{1}^{\top} \bs{v}\}.
\end{equation}
Then we have the following inequalities:
\begin{subequations}
\begin{align}
|det(\bs{A})| & \leq \prod_{i=1}^{C}\|\bs{A}_{:, i}\|_{2} \label{sub2}\\
& \leq \prod_{i=1}^{C} \mathbf{1}^{\top}\bs{A}_{:, i} \label{sub3}\\
& \leq(\frac{\sum_{i=1}^{C} \mathbf{1}^{\top}\bs{A}_{:, i}}{C})^{C} \label{sub4}\\
&=(\frac{\mathbf{1}^{\top} \bs{A} \mathbf{1}}{C})^{C}=1, \label{sub5}
\end{align}
\label{15}
\end{subequations}

where (\ref{sub2}) is Hadamard's inequality; (\ref{sub3}) is by Eq. (\ref{13}); (\ref{sub4}) is by the arithmetic-geometric mean inequality; and (\ref{sub5}) is by Eq. (\ref{eq.11}). 

Note that $|det(\bs{A})|^{-1}=|det(\bs{A}^{-\top})|$ and $det(\bs{T}_{\star})=det( \bs{T} \bs{A}^{-\top})=det(\bs{T})|det(\bs{A})|^{-1}$ from properties of the determinant, it follows from Eq.~(\ref{15}) that $det(\bs{T}_{\star}) \geq det( \bs{T})$. We also know that $det(\bs{T}_{\star}) \leq det( \bs{T})$ must hold from Criterion (5), hence we have 
\begin{equation}
	det(\bs{T}_{\star}) = det( \bs{T})
\end{equation}

By Hadamard's inequality, the equality in (\ref{sub2}) holds only if $\bs{A}$ is column-orthogonal, which is equivalent to that $\bs{A}^{-\top}$ is column-orthogonal. Considering condition (2) in the definition of sufficiently scattered and the property of $\bs{A}^{-\top}$ that $\mathrm{cone}\{\bs{H}\} \subseteq \mathrm{cone}(\bs{A}^{-\top})$, the only possible choices of column-orthogonal $\bs{A}^{-\top}$ are

\begin{equation}
\bs{A}^{-\top}=\boldsymbol{\Pi} \boldsymbol{\Phi}
\end{equation}

where $\boldsymbol{\Pi} \in \mathbb{R}^{C \times C}$ is any permutation matrix and $\boldsymbol{\Phi} \in \mathbb{R}^{C \times C}$
is any diagonal matrix with non-zero diagonals. By Eq. (\ref{eq.11}), we must have $\Phi=$ I. Subsequently, we are left with $\bs{A}^{-\top}=\bs{\Pi}$, or equivalently, $\bs{T}_{\star} = \bs{\Pi} \bs{T}.$ Since $\bs{T}$ and $\bs{T}_{\star}$ are both diagonal dominant, the only possible permutation matrix is $\bs{I}$, which means $\bs{T}_{\star} = \bs{T}$ holds. By Eq. (\ref{eq.8}), it follows that $h_{\bs{\theta}_\star}= \clpos$. Hence we conclude that  $(\bs{T}_{\star}, h_{\bs{\theta}_\star})=(\bs{T},\clpos)$ is the unique optimal solution to criterion (5).  \qed

  \begin{figure*}
    
     \includegraphics[width=0.24\textwidth]{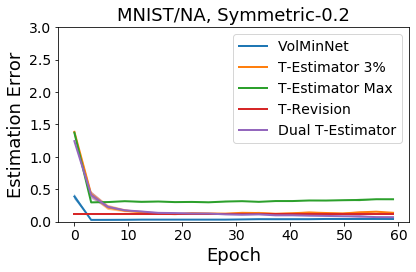}
     \includegraphics[width=0.24\textwidth]{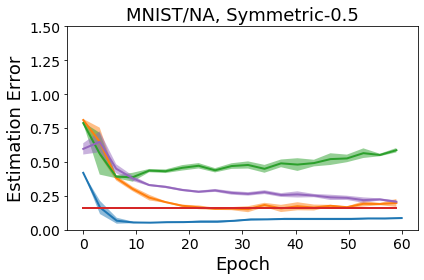}   
     \includegraphics[width=0.24\textwidth]{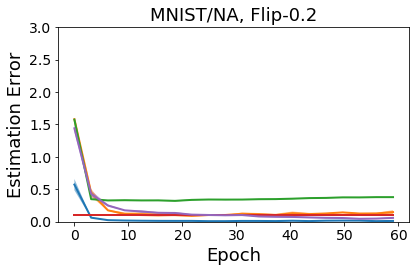}
     \includegraphics[width=0.24\textwidth]{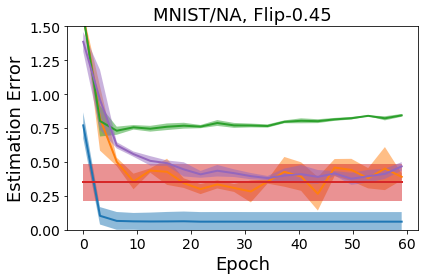}
     \vspace{5mm}

     \includegraphics[width=0.24\textwidth]{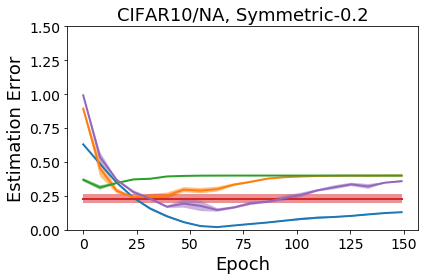}
     \includegraphics[width=0.24\textwidth]{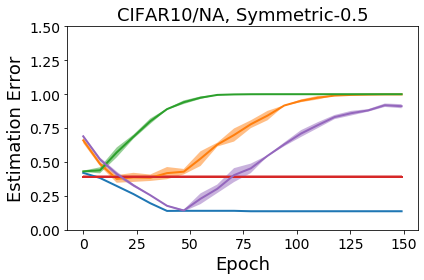}
     \includegraphics[width=0.24\textwidth]{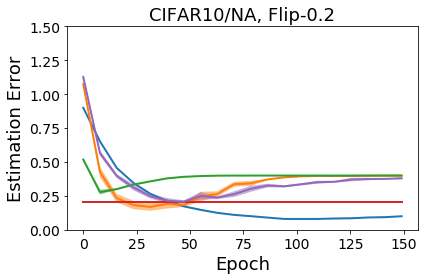}
     \includegraphics[width=0.24\textwidth]{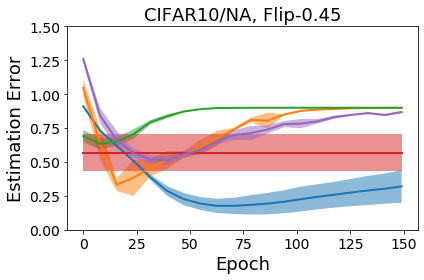}
     
     \vspace{5mm}

     \includegraphics[width=0.24\textwidth]{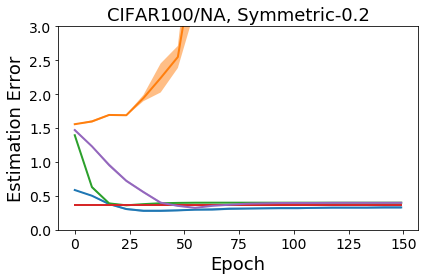}
     \includegraphics[width=0.24\textwidth]{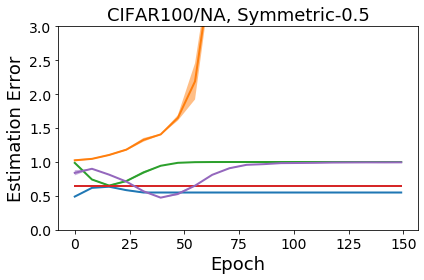}
     \includegraphics[width=0.24\textwidth]{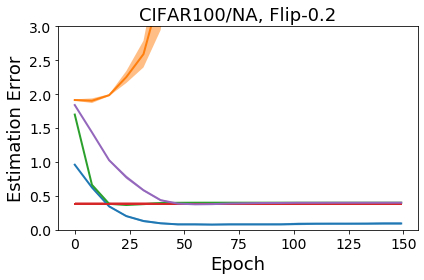}
     \includegraphics[width=0.24\textwidth]{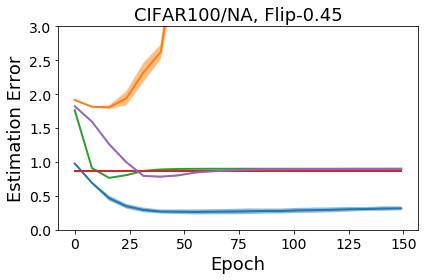}
    
     \caption{Transition matrix estimation error on MNIST/NA, CIFAR-10/NA, CIFAR-100/NA. Datasets with ``/NA'' means that possible anchor points are removed. The error bar for the standard deviation in each figure has been shaded. The lower the better.}
     \label{fig:na_est_error}
 \end{figure*}

\section{Experiments on datasets where possible anchor points are manually removed.}

 \begin{table*}

    \centering
    \scalebox{1}{
\begin{tabular}{ccccccc}

\hline  
& \multicolumn{2}{c} {MNIST/NA} & \multicolumn{2}{c} {CIFAR-10/NA} & \multicolumn{2}{c} {CIFAR-100/NA} \\
& Sym-20\% & Sym-50\% & Sym-20\% & Sym-50\% & Sym-20\% & Sym-50\% \\
\hline
Decoupling & $ 96.72 \pm 0.16$ & $ 92.72 \pm 0.33$ & $ 75.51 \pm 0.38$ & $ 49.96 \pm 0.51$ & $ 38.83 \pm 0.37$ & $ 20.42 \pm 0.53$ \\
MentorNet & $ 97.10 \pm 0.10$ & $ 95.10 \pm 0.14$ & $ 80.25 \pm 0.52$ & $ 71.65 \pm 0.28$ & $ 39.72 \pm 0.35$ & $ 29.39 \pm 0.35$ \\
Co-teaching & $ 97.06 \pm 0.12$ & $ 94.89 \pm 0.10$ & $ 81.74 \pm 0.32$ & $ 73.38 \pm 0.45$ & $ 44.92 \pm 0.11$ & $ 33.13 \pm 0.88$\\
Forward & $ 98.46 \pm 0.07$ & $ 97.59 \pm 0.05$ & $ 84.25 \pm 0.22$ & $ 70.00 \pm 3.07$ & $ 50.58 \pm 0.68$ & $ 36.79 \pm 1.86$  \\
T-Revision & $ 98.72 \pm 0.13$ & $ 97.86 \pm 0.11$ & $ 86.81 \pm 0.19$ & $ 74.10 \pm 2.34$ & $ 59.57 \pm 1.13$ & $ 43.75 \pm 0.84$ \\
DMI  & $ 98.42 \pm 0.03$ & $ 97.87 \pm 0.18$ & $ 83.42 \pm 0.54$ & $ 77.82 \pm 0.45$ & $ 56.29 \pm 0.28$ & $ 41.81 \pm 0.70$\\
Dual T & $ 98.61 \pm 0.12$ & $ 97.91 \pm 0.12$ & $ 86.70 \pm 0.06$ & $ 78.92 \pm 0.42$ & $ 56.99 \pm 1.00$ & $ 42.04 \pm 1.96$\\
VolMinNet & $ \mathbf{98.72} \pm \mathbf{0.06}$ & $ \mathbf{97.94} \pm \mathbf{0.07}$ & $ \mathbf{88.72} \pm \mathbf{0.03}$ & $ \mathbf{82.38} \pm \mathbf{0.65}$ & $ \mathbf{63.40} \pm \mathbf{1.25}$ & $ \mathbf{51.04} \pm \mathbf{1.23}$ \\
\hline
\end{tabular}
}
    \centering
    \scalebox{1}{
\begin{tabular}{ccccccc}

&\multicolumn{2}{c} {MNIST/NA}  &\multicolumn{2}{c} {CIFAR-10/NA} & \multicolumn{2}{c} {CIFAR-100/NA} \\
& Pair-20\% & Pair-45\% &Pair-20\% & Pair-45\% & Pair-20\%& Pair-45\% \\
\hline
Decoupling & $ 96.92 \pm 0.06$ & $ 93.29 \pm 0.57$ & $ 77.06 \pm 0.26$ & $ 50.81 \pm 0.73$ & $ 40.42 \pm 0.47$ & $ 26.21 \pm 0.67$ \\
MentorNet & $ 96.88 \pm 0.04$ & $ 88.17 \pm 0.70$ & $ 77.62 \pm 0.28$ & $ 57.60 \pm 0.35$ & $ 39.11 \pm 0.41$ & $ 25.17 \pm 0.36$\\
Co-teaching & $ 96.96 \pm 0.07$ & $ 95.34 \pm 0.09$ & $ 80.70 \pm 0.18$ & $ 69.15 \pm 0.89$ & $ 43.04 \pm 0.73$ & $ 26.67 \pm 0.29$ \\
Forward & $ 98.61 \pm 0.33$ & $ 78.51 \pm 17.48$ & $ 85.87 \pm 0.82$ & $ 53.92 \pm 11.39$ & $ 51.37 \pm 0.99$ & $ 34.69 \pm 1.37$ \\
T-Revision & $ 98.71 \pm 0.31$ & $ 82.65 \pm 14.61$ & $ 87.52 \pm 0.58$ & $ 53.96 \pm 14.67$ & $ 59.70 \pm 1.43$ & $ 38.35 \pm 0.60$\\
DMI  & $ 98.78 \pm 0.11$ & $ 97.46 \pm 1.38$ & $ 86.14 \pm 1.52$ & $ 70.01 \pm 5.63$ & $ 54.05 \pm 1.09$ & $ 35.03 \pm 2.91$\\
Dual T & $ 98.76 \pm 0.13$ & $ 85.77 \pm 7.85$ & $ 89.02 \pm 0.40$ & $ 65.17 \pm 0.72$ & $ 59.07 \pm 3.79$ & $ 36.95 \pm 3.19$\\
VolMinNet & $ \mathbf{98.87} \pm \mathbf{0.11}$ & $ \mathbf{97.80} \pm \mathbf{2.43}$ & $ \mathbf{89.26} \pm \mathbf{0.22}$ & $ \mathbf{84.48} \pm \mathbf{3.85}$ & $ \mathbf{64.88} \pm \mathbf{1.87}$ & $ \mathbf{56.07} \pm \mathbf{3.35}$  \\
\hline
\end{tabular}
}

	\caption{Classification accuracy (percentage) on MNIST, CIFAR-10,CIFAR-100 and MNIST/NA, CIFAR-10/NA, CIFAR-100/NA. Datasets with ``/NA'' means that possible anchor points are removed.}
	\label{table:na_accs}
\end{table*}

Following Xia et al.\yrcite{xia2019anchor}, to show the importance of anchor points, we remove possible anchor points from the datasets, i.e., instances with large estimated class-posterior probability $P(Y | X),$ before corrupting the training and validation sets. For MNIST we removed $40\%$ of the instances with the largest estimated class posterior probabilities in each class. For CIFAR-10 and CIFAR-100, we removed $10 \%$ of the instances with the largest estimated class posterior probabilities in each class. We add "/NA" following the dataset's name denote those datasets which are modified by removing possible anchor points. The detailed experimental results are shown in Figure \ref{fig:na_est_error} (estimation error) and Table \ref{table:na_accs} (classification accuracy). The experimental performance shows that our proposed method outperforms the baseline methods.

%
%
%

\end{document}